\documentclass[letterpaper]{article} 
\usepackage{aaai24}  
\usepackage{times}  
\usepackage{helvet}  
\usepackage{courier}  
\usepackage[hyphens]{url}  
\usepackage{graphicx} 
\usepackage{natbib}  
\usepackage{caption} 
\usepackage{algorithm}
\usepackage{algorithmic}

\usepackage{newfloat}
\usepackage{listings}
\DeclareCaptionStyle{ruled}{labelfont=normalfont,labelsep=colon,strut=off} 
\floatstyle{ruled}
\newfloat{listing}{tb}{lst}{}
\floatname{listing}{Listing}
\usepackage{color}
\definecolor{applegreen}{rgb}{0.55, 0.71, 0.0}

\usepackage{siunitx}
\usepackage{amsmath}

\usepackage{subfigure,graphicx}
\usepackage{pgfplots}
\usepackage{multirow}
\usepackage{booktabs}

\usepackage{etoolbox}
\makeatletter
\let\@titlehook=\relax
\apptocmd{\@maketitle}{\@titlehook}{}{}
\newcommand{\titlehook}[1]{\def\@titlehook{#1}}
\makeatother

\title{Enhancing the Authenticity of Rendered Portraits with Identity-Consistent Transfer Learning}
\author {
    Luyuan Wang\equalcontrib\textsuperscript{\rm 1},
    Yiqian Wu\equalcontrib\textsuperscript{\rm 1},
    Yongliang Yang\textsuperscript{\rm 2},
    Chen Liu\textsuperscript{\rm 3},
    Xiaogang Jin\thanks{Corresponding author.}\textsuperscript{\rm 1}
}
\affiliations {

    \textsuperscript{\rm 1}State Key Lab of CAD\&CG, Zhejiang University, Hangzhou 310058, China\\
    \textsuperscript{\rm 2}Department of Computer Science, University of Bath, UK\\
    \textsuperscript{\rm 3}Zhejiang Linctex Digital Technology Co. Ltd., China\\

}

\usepackage{bibentry}

\pgfplotsset{compat=1.16}
\begin{document}



\titlehook{
    \begin{center}
        \centering
        \captionsetup{type=figure}
        \includegraphics[width=1.0\textwidth]{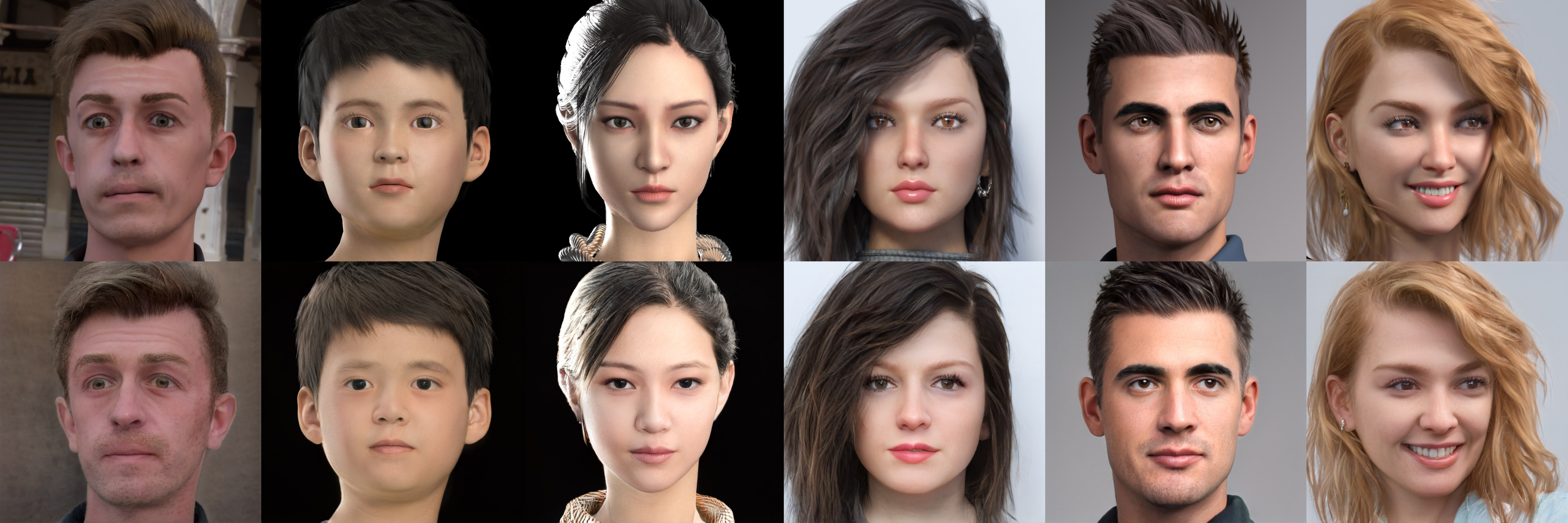}
        \captionof{figure}{
        {Given rendered 3D faces as input (top row), our method effectively mitigates the ``uncanny valley'' effect and improves the overall authenticity of rendered portraits while preserving facial identity (bottom row).}
          Please zoom in for a better view.
        }
        \label{fig:teaser}
    \end{center}%
}
\nocopyright
\maketitle


\begin{abstract}
Despite rapid advances in computer graphics, creating high-quality photo-realistic virtual portraits is prohibitively expensive. Furthermore, the well-known ``uncanny valley'' effect in rendered portraits has a significant impact on the user experience, especially when the depiction closely resembles a human likeness, where any minor artifacts can evoke feelings of eeriness and repulsiveness. In this paper, we present a novel photo-realistic portrait generation framework that can effectively mitigate the ``uncanny valley'' effect and improve the overall authenticity of rendered portraits. 
Our key idea is to employ transfer learning to learn an identity-consistent mapping from the latent space of rendered portraits to that of real portraits.
During the inference stage, the input portrait of an avatar can be directly transferred to a realistic portrait by changing its appearance style while maintaining the facial identity. 
To this end, we collect a new dataset, \textbf{D}az-\textbf{R}endered-\textbf{F}aces-\textbf{HQ} (\textit{DRFHQ}), that is specifically designed for rendering-style portraits. We leverage this dataset to fine-tune the StyleGAN2 generator, using our carefully crafted framework, which helps to preserve the geometric and color features relevant to facial identity.
We evaluate our framework using portraits with diverse gender, age, and race variations. 
Qualitative and quantitative evaluations and ablation studies show the advantages of our method compared to state-of-the-art approaches.

\end{abstract}

\section{Introduction}
\label{sec::Introduction}
Generating photo-realistic and indistinguishable faces from 3D renderings has long been a challenge.
Over the last two decades, the growth of entertainment industries such as animation, film, and video games has led to tremendous advances in high-quality face modeling and rendering technology.
Under passive illumination, approaches based on multi-view stereo systems \cite{10.1145/1833349.1778777, 10.1145/2010324.1964970, Bradley2010, DBLP:journals/cgf/FyffeNHSBJLD17, Riviere2020} can reconstruct 
high-quality face geometry. Following the pioneering work of Debevec et al. \shortcite{Debevec2000}, 
a series of light-stage-based facial appearance capture methods \cite{sun2020light, Ghosh2011, Fyffe2016, Gotardo2015} have been proposed to capture the pore-level properties of a human face.
While these approaches can help produce high-quality faces, they are highly expensive and time-consuming. 
Furthermore, even though these faces are of superior quality, they contain subtle unrealistic details that are immediately noticeable because humans are innately sensitive to such details when perceiving faces.
The ``uncanny valley'' effect, first described by Japanese roboticist Masahiro Mori \cite{mori1970bukimi}, shows how imperfectly human-like objects, such as robots, 3D animations, and life-like dolls, can have a negative impact on user experience and interaction  \cite{moore2012bayesian}. Because these impersonations do not have a lifelike appearance, they can cause a sudden shift in a person's response from empathy to eerie, frightening, or revulsion, also known as ``uncanny" sensations.

The rise of deep learning, in particular Generative Adversarial Networks (GANs) \cite{DBLP:conf/nips/GoodfellowPMXWOCB14}, has inspired researchers to develop high-quality face generation methods \cite{Zhu_2017_ICCV, Choi_2020_CVPR}.
In recent years, StyleGAN \cite{DBLP:conf/cvpr/KarrasLA19} and its variants \cite{DBLP:conf/cvpr/KarrasLAHLA20,DBLP:conf/nips/KarrasAHLLA20,DBLP:conf/nips/KarrasALHHLA21}
have paved the way for the semantic manipulation of photo-realistic portraits.
The existing methods that can improve the realism of avatar faces \cite{DBLP:conf/eccv/GarbinKJS20,DBLP:journals/tog/ChandranWZRGGB21} are all based on projecting the rendering-style faces into the pretrained StyleGAN2 generator, thanks to its high generation quality and diversity.
Garbin et al. \shortcite{DBLP:conf/eccv/GarbinKJS20} 
matches a non-photorealistic portrait to a latent code of the pretrained StyleGAN2 generator while maintaining pose, expression, hair, and lighting consistency. 
Despite the attempt to adapt to the real face domain, their method necessitates intricate and time-consuming processing.
Furthermore, since the input is out of the domain of the pretrained model, the output often has artifacts such as distortion and identity inconsistency. 
Chandran et al. \shortcite{DBLP:journals/tog/ChandranWZRGGB21} project high-quality yet incompletely rendered facial skin into the latent space of StyleGAN2, 
generating temporally-coherent and photo-realistic portraits. 
Nevertheless, their method is more of an inpainting process for the missing face components, such as hair, eyes, and mouth interior. 
Also, the output images still retain the rendering style thus lack authenticity.

The limitations of the existing works motivate us to present a novel StyleGAN-based portrait generation framework to increase the authenticity of 
rendered portraits.
We propose a transfer-learning-based approach to establish the correlation between portrait images with different styles.
The key idea is to develop an identity-consistent fine-tuning method that results in a rendering-style generator with facial identities matching those of the realistic-style StyleGAN2 generator.
We treat a latent code in the $W+$ latent space of a portrait as an implicit representation of both portrait style and identity.
While the portrait style can be either a rendering style or a realistic style corresponding to the two generators, the portrait identity is shared in-between.
That is, if we project a rendered portrait into the rendering-style generator's {$W+$} latent space, the realistic-style StyleGAN2 generator can interpret the resulting latent code as a realistic portrait with the rendered portrait's facial identity. 
We find that by doing so, the rendering-style can be effectively removed from the final output, and the facial identity can be preserved without distortion.
Based on this principle, we first collect a new dataset of rendering-style portraits, \textbf{D}az-\textbf{R}endered-\textbf{F}aces-\textbf{HQ} (\textit{DRFHQ}). 
Inspired by StyleGAN2-ada \cite{DBLP:conf/nips/KarrasAHLLA20}, we use \textit{DRFHQ} to finetune the StyleGAN2-\textit{FFHQ} generator, which has been pretrained on the \textit{FFHQ} dataset, resulting in a rendering-style StyleGAN2-\textit{DRFHQ} generator.
During finetuning, we constrain with sketches and color to help the new generator maintain facial identities.
Then we perform latent code optimization to project the input rendering-style portrait into StyleGAN2-\textit{DRFHQ}'s latent space. 
Finally, we feed the resulting latent code into the pretrained StyleGAN2-\textit{FFHQ} generator, yielding a photo-realistic portrait with preserved 
facial identity.
Extensive evaluations demonstrate that our work is capable of generating plausible results for rendered portraits.

In summary, our work makes the following contributions:
\begin{itemize}
    \item{
    We present a novel portrait generation framework to overcome the ``uncanny valley'' effect for rendered 3D faces.   
    }
         \item{
             Based on a new high-quality rendering-style portrait dataset (\textit{DRFHQ}), we propose a novel transfer learning approach to correlate portraits with different styles in the learned latent space while preserving facial identity. 
         }
\end{itemize}

\section{Related Work}
\label{sec::Related Work}

\paragraph{Portrait Synthesis}
Human face modeling and rendering is a crucial and active research topic for applications in the entertainment, film, and television industries.
Most physically-based rendering methods require a multi-view stereo system to reconstruct pore-level geometry and skin reflectance properties \cite{DBLP:journals/cgf/FyffeNHSBJLD17,10.1145/3550454.3555509,DBLP:conf/cvpr/LiBZCIXRPKXL20,10.1145/2010324.1964970,Riviere2020}.
To capture detailed human faces, 
a number of light-stage-based approaches \cite{sun2020light, Ghosh2011, Fyffe2016, Gotardo2015} have been developed based on the seminal work for facial appearance capturing and reconstruction \cite{Debevec2000}.
Although the photo-realistic renderings of avatars are almost indistinguishable from real humans,
the ``uncanny valley'' effect occurs when an anomaly is revealed from their seemingly realistic appearance \cite{DBLP:journals/presence/SeyamaN07}.
Researchers have suggested methods to measure the ``uncanny valley'' effect \cite{DBLP:journals/presence/SeyamaN07,DBLP:journals/ijsr/HoM17},
however, it is difficult to eliminate such an unpleasant effect in traditional rendering.
Since their introduction in 2020, neural radiance fields (NeRF) \cite{DBLP:conf/eccv/MildenhallSTBRN20} have spawned a slew of downstream applications, including face synthesis \cite{10.1145/3550454.3555501,Hong_2022_CVPR,10.1007/978-3-031-20062-5_16}.
Researchers also combine NeRF with generators \cite{DBLP:conf/nips/SchwarzLN020,DBLP:conf/iclr/GuL0T22,DBLP:conf/cvpr/ChanMK0W21,DBLP:conf/cvpr/XueLSL22} to support view-consistent image synthesis without the need for multi-view images of a specific person.
However, existing face modelling and rendering methods still struggle to produce photo-realistic results that can avoid the ``uncanny valley'' effect.
The introduction of generative adversarial networks (GANs)~\cite{DBLP:conf/nips/GoodfellowPMXWOCB14} sparks an increasing number of face synthesis models \cite{DBLP:conf/nips/GulrajaniAADC17,DBLP:conf/iclr/KarrasALL18}.
Among these works, StyleGAN \cite{DBLP:conf/cvpr/KarrasLA19,DBLP:conf/cvpr/KarrasLAHLA20,DBLP:conf/nips/KarrasALHHLA21} is mostly favored due to its synthesis quality and manipulation ability, and serves as an inspiration for many downstream works \cite{DBLP:conf/mm/PariharDKR22,DBLP:conf/mm/ZhangBG21}.
We also build our framework based on the StyleGAN model since it not only provides important prior information of facial identity and appearance on various human faces, but also allows efficient portrait editing at a high level.

\paragraph{Face Style Transfer using StyleGAN}
Portrait style transfer using StyleGAN is also related to our work.
Pinkney and Adler \shortcite{DBLP:journals/corr/abs-2010-05334} use a resolution-dependent method to interpolate different styles at appearance and geometry levels.
StyleCariGAN \cite{DBLP:journals/tog/JangJJYT021} modulates coarse layer feature maps of StyleGAN by shape exaggeration blocks to produce desirable caricature shape exaggerations.
However, it requires a dataset that contains thousands of images,
whereas other approaches have been proposed to reduce the dataset size to a few hundred \cite{Yang_2022_CVPR}, $\sim$ 100 \cite{DBLP:journals/tog/SongLLMLZC21,DBLP:journals/tog/MenYCLX22}, $\sim$ 10 \cite{Ojha_2021_CVPR}, or even to achieve one-shot domain adaptation \cite{DBLP:conf/iclr/ZhuA0W22, Zhang2022GeneralizedOD}.
Wu et al. \shortcite{wu2021stylealign} conduct a thorough investigation into the properties of aligned StyleGAN and use their findings to investigate potential applications such as cross-domain image morphing and zero-shot vision tasks.
In addition to example images, StyleGAN-NADA \cite{DBLP:journals/tog/GalPMBCC22} uses text prompt as input to stylize portraits with the help of a pretrained CLIP model.
This line of research has been expanded to videos \cite{10.1145/3550454.3555437} to achieve consistent results in a sequence.
Sang et al. \shortcite{DBLP:conf/siggrapha/SangZS0LLW0L22} also attempt to create stylized and editable 3D models directly from users' avatars.
However, the above methods are intended to generate stylized portraits from real photos, whereas our work aims at the opposite: transfer the ``rendering-style'' of the rendered portraits into the ``realistic-style'' of the results that are indistinguishable from real portraits.

\paragraph{Face Realism Improvement using StyleGAN}
Improving the realism of rendered faces is still a challenging issue.
Garbin et al. \shortcite{DBLP:conf/eccv/GarbinKJS20} propose a zero-shot image projection algorithm that requires no training data to find the latent code that most closely matches the input rendered face. Their objective is the most similar to ours. However, their method requires a significant amount of processing time and may result in inconsistencies in facial identity. 
Chandran at al. \shortcite{DBLP:journals/tog/ChandranWZRGGB21} use a multi-frame consistent method to project the traditional incomplete face rendering results into latent space to achieve photo-realistic rendering and animation of a full-head portrait.
Despite generating realistic full-head portraits, their primary goal is to inpaint the missing components. As a result, their method preserves the input rendered skin but is incapable of improving the authenticity of resultant faces.
The StyleGAN encoders \cite{DBLP:journals/tog/TovANPC21,DBLP:conf/cvpr/RichardsonAPNAS21,10.1145/3544777,DBLP:conf/iccv/AlalufPC21,DBLP:conf/cvpr/AlalufTMGB22} and some optimization-based methods \cite{Abdal_2019_ICCV,10.1145/3544777,Abdal_2020_CVPR} can project the rendered faces into StyleGAN's latent space. However, the rendered faces are far outside the domain of the real faces, thus resulting in distortion and artifacts or maintaining the ``rendering-style''.
Different from these methods, we focus on producing realistic portraits for digital 3D faces
while preserving the facial identity.

\begin{figure}[t]
    \centering
    \includegraphics[width=\linewidth]{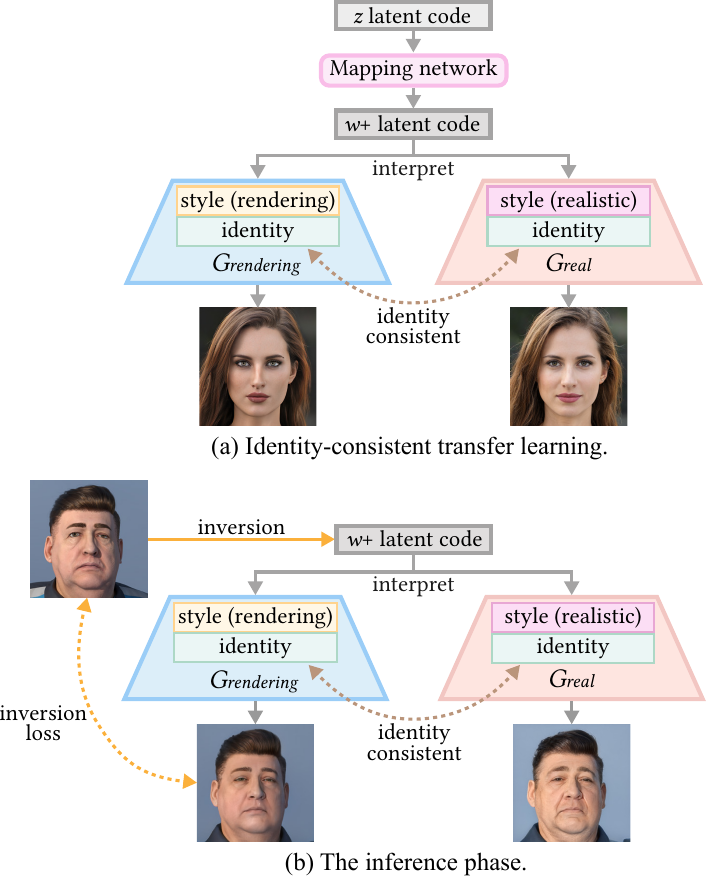}
    \caption{\label{fig:motivation}The central idea of our method. 
(a) In identity-consistent transfer learning, a single latent code in the $W+$ latent space can be interpreted as portraits with the same facial identity but different image styles by $G_{real}$ and $G_{rendering}$.
(b) In our inference phase, we invert the input rendered portrait to the $W+$ latent space of $G_{rendering}$. The resulting latent codes can be interpreted as a realistic-style portrait by $G_{real}$ while preserving the facial identity of the input rendered portrait. The rendered portrait is from the \textit{Diverse Human Faces} \cite{DHFdataset} dataset.}
\end{figure}

\begin{figure*}[t]
    \centering
    \includegraphics[width=0.94\linewidth]{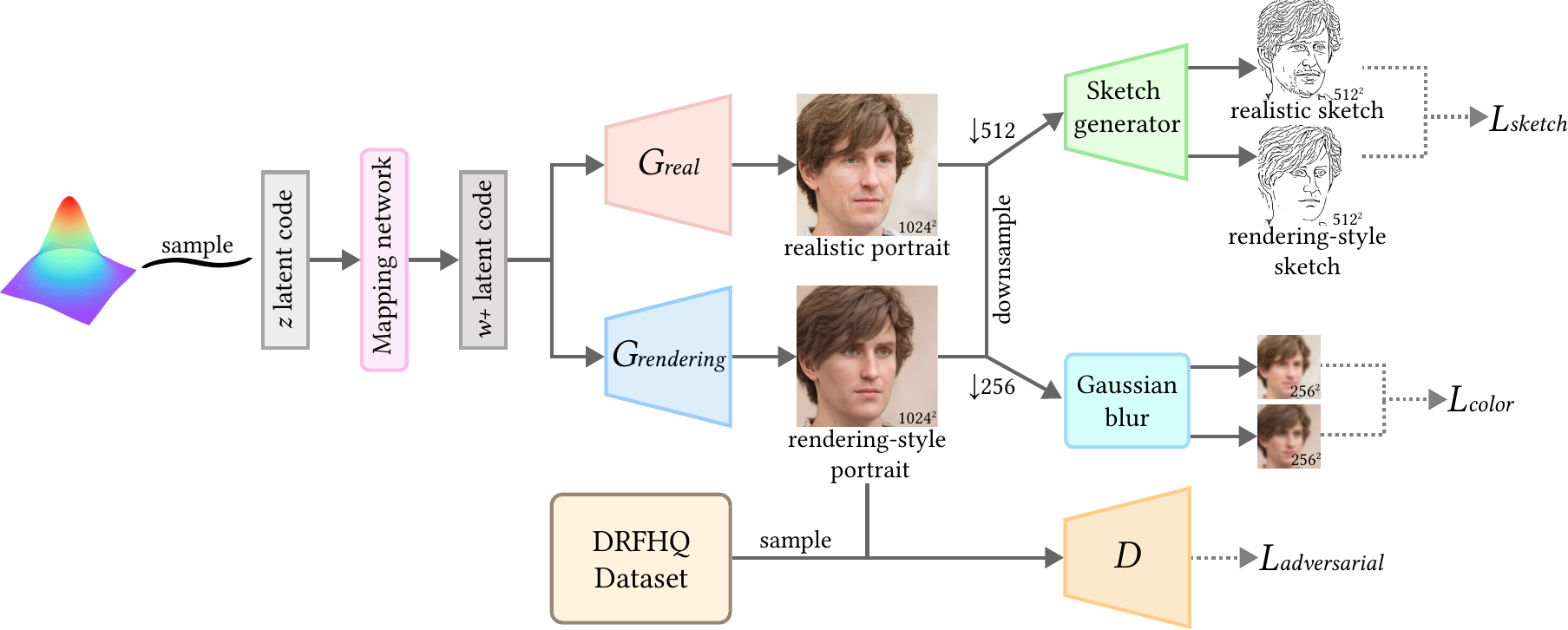}
    \caption{\label{fig:net_pipeline}An overview of our identity-consistent transfer learning network's training stage. We begin by initializing a rendering style generator $G_{rendering}$ with the weights of the pretrained StyleGAN2-\textit{FFHQ} generator ($G_{real}$), which can provide various face priors. During the finetuning of the generator, 
    we only update $G_{rendering}$ and the discriminator, while the sketch generator, $G_{real}$, and the mapping network are frozen.
    The $L_{color}$ and $L_{sketch}$ loss encourages small variations in color and face contour between the image generated by $G_{real}$ and the image generated by $G_{rendering}$. 
    }
\end{figure*}

\section{Method}
\label{sec:Method}
Our objective is to improve the authenticity of digital 3D faces by substituting them with photo-realistic versions that are indistinguishable, all while preserving the avatar's inherent facial identity.
Fig. \ref{fig:motivation} demonstrates the key idea of our approach.
We conduct portrait replacement in the latent space by employing latent code that implicitly represents portrait style and identity as the interface in-between.
As shown in Fig. \ref{fig:motivation} (a), we establish identity-consistent transfer learning on the StyleGAN generator of realistic portraits ($G_{real}$), resulting in a finetuned generator ($G_{rendering}$) of portraits with a different style, i.e., the ``rendering'' style.
The transfer learning is performed in a way that given a single latent code in the $W+$ latent space, the portrait identity can be well preserved in both generators, only the portrait style is interpreted differently as ``realistic-style'' by $G_{real}$ and ``rendering-style'' by $G_{rendering}$. 
In other words, the same latent code can generate two portraits with distinct styles but matched identity. 
Unlike the style change in the finetuning process, during the inference phase (see Fig. \ref{fig:motivation} (b)), we aim to ``invert'' the style of the input portrait from ``rendering'' to ``realistic''. 
We begin by applying GAN inversion to obtain the avatar's latent code in the $W+$ latent space of $G_{rendering}$. The latent code is then fed into $G_{real}$ to adapt to the realistic style while preserving identity. 
In the end, we achieve the final result - a photo-realistic portrait with the identity of the input avatar.

The rest of the section is organized as follows.
We begin by introducing \textit{DRFHQ}, a new high-quality rendering-style portrait dataset used for transfer learning (Sec. \ref{sec:Daz-Rendered-Faces-HQ dataset}). 
Then we elaborate our transfer learning strategy, which finetunes $G_{real}$ to achieve $G_{rendering}$ with a different style 
while minimizing other irrelevant changes (Sec. \ref{sec:Identity-Consistent Transfer Learning}).
Finally, we present how we increase the authenticity of rendered portraits
in the inference phase (Sec. \ref{sec:Inference}).

\subsection{Daz-Rendered-Faces-HQ dataset}
\label{sec:Daz-Rendered-Faces-HQ dataset}
We create \textbf{D}az-\textbf{R}endered-\textbf{F}aces-\textbf{HQ} (\textit{DRFHQ}), a dataset that comprises high-quality rendering-style portrait images, by collecting daz3d.com's gallery \cite{Daz3dWebsite}. 
\textit{DRFHQ} contains 11,399 high-quality PNG images in $1024\times1024$ resolution, with a wide range of gender, age, pose, race, hairstyle, etc.
We first align and crop the raw images using Dlib \cite{DBLP:conf/cvpr/KazemiS14} according to the preprocessing method of \textit{FFHQ}, then manually filter the aligned images.
Due to copyright restrictions, we cannot release the collected images but will provide the corresponding URLs as an alternative. 
Although several publicly available rendering-style datasets exist \cite{wood2021fake,DBLP:conf/nips/LiuLQZWZ21,mascaro2020uibvfed,DHFdataset}, 
their face resolution is insufficient for high-quality digital face display \cite{wood2021fake,DHFdataset}, or they only contain a small number of rendered faces \cite{DBLP:conf/nips/LiuLQZWZ21,mascaro2020uibvfed}, or they are rendered using a small number of face models (100 different identities) \cite{DHFdataset}.
\textit{DRFHQ} is the first high-quality rendering-style dataset with a face region resolution of $1024\times1024$ that can be extended to downstream tasks, to the best of our knowledge.

\subsection{Identity-Consistent Transfer Learning}
\label{sec:Identity-Consistent Transfer Learning}
Inspired by StyleGAN-ada \cite{DBLP:conf/nips/KarrasAHLLA20}, we use \textit{DRFHQ} to finetune the pretrained StyleGAN2-\textit{FFHQ} generator $G_{real}$, resulting in a new stylized generator $G_{rendering}$ capable of producing rendering-style portraits. 
However, simply finetuning $G_{real}$ 
leads to large facial identity deviations in the finetuned latent space compared to the original.
To address this issue, we use two additional losses during the finetuning (training) process to constrain the facial identity. The training pipeline is illustrated in Fig. \ref{fig:net_pipeline}. 

Our idea is to use the same latent code in $W+$ latent space to implicitly represent the rendered face and its realistic face replacement, hence $G_{rendering}$ and $G_{real}$ are required to share the same $W+$ latent space. To do this, we freeze the mapping network during finetuning, resulting in a single latent code $z$ in $Z$ latent space being mapped to the same latent code 
$w+ \in W+$ of $G_{rendering}$ and $G_{real}$. We will omit the unmodified mapping network in the remainder of this section and use $w+$ as the latent code.

\textbf{Sketch loss.} Inspired by DeepFaceEditing \cite{Chen2021}, the geometric features of the face can be well represented by sketches. Therefore, we add the following L1 loss function:
\begin{equation}
	\begin{split}
    \label{equation-sketch-loss}
     L_{sketch} = & \Vert S(G_{real}(w+)\downarrow_{512}) \\
                  &-S(G_{rendering}(w+)\downarrow_{512})\Vert_1,
     \end{split}
\end{equation}
where $G_{rendering}$ is to be finetuned and initialized by the pretrained $G_{real}$, $S$ is the pretrained sketch extractor in DeepFaceEditing \cite{Chen2021} model, and $\downarrow_{512}$ denotes the interpolation operation that downsamples the
images to $512\times512$.
According to Eq.~\ref{equation-sketch-loss}, the output of $G_{real}$ and $G_{rendering}$ are fed into $S$ separately to obtain two face sketches, and the geometric contours of the two faces are constrained to be as similar as possible by using the L1 norm.

\textbf{Color loss.} To preserve the portrait color during transfer learning, we propose a color loss at the perceptual level based on the LPIPS loss \cite{DBLP:conf/cvpr/ZhangIESW18}.
However, LPIPS captures the facial appearance similarity, including texture and style-related details, preventing the generator from learning rendering-style. 
Inspired by \cite{DBLP:conf/eccv/GarbinKJS20}, we solve this problem by removing the appearance details from the images. Specifically, we first downsample the images to $256 \times 256$ and apply Gaussian blur, then feed the images into the VGG16 network to compute the LPIPS loss:
%
\begin{equation}
	\begin{split}
\label{equation-color-loss}
     L_{color} = LPIPS(&B(G_{real}(w+)\downarrow_{256}), \\
                       &B(G_{rendering}(w+)\downarrow_{256}),
     \end{split}
\end{equation}
where $B$ is the Gaussian blur operation with $kernel=13$ and $\sigma=10$, and  $\downarrow_{256}$ denotes the interpolation operation that downsamples the images
to $256\times256$.

Our objective loss function used in finetuning is the weighted sum of the above two losses:
\begin{equation}
\label{equation-generator-overall-loss}
    L_{G} = \lambda_s L_{sketch} + \lambda_c L_{color},
\end{equation}
where we empirically set $\lambda_s = 5\times 10^{-6}$ and $\lambda_c = 3.75\times 10^{3}$. 
$L_{G}$ is added to the original loss of StyleGAN2-ada to finetune the generator.

\subsection{Inference}
\label{sec:Inference}
In the inference phase, we use a direct latent optimization \cite{DBLP:conf/cvpr/KarrasLAHLA20} inversion method
to project the rendered portrait $x$ into the latent space of $G_{rendering}$.
As we aim for the least distortion instead of the best editability, 
we optimize in the $W+$ latent space, which has greater expressive potential:
\begin{equation}
	\begin{split}
\label{equation-inversion}
     w+^*,n^* = & \mathop{\arg\min}\limits_{w+,n} LPIPS(x, G_{rendering}(w+,n)) \\
                & + \lambda_n L_n(n),
     \end{split}
\end{equation}
where $G_{rendering}(w+,n)$ is image generated by $G_{rendering}$ with noise $n$, $L_n$ is a noise regularization term, and $\lambda_n = 1e5$. We initialize $w+$ as the average latent code in the $W+$ latent space and use a 500-step optimization to get $w+^*$.
Finally, we input the resulting latent code $w+^*$ to $G_{real}$, yielding a photo-realistic portrait. We do not employ the optimized noise $n^*$ here because the regularization term $L_n$ prevents the noise vector from influencing the final result.


\begin{figure}[htbp] 
\centering
\includegraphics[width=\linewidth]{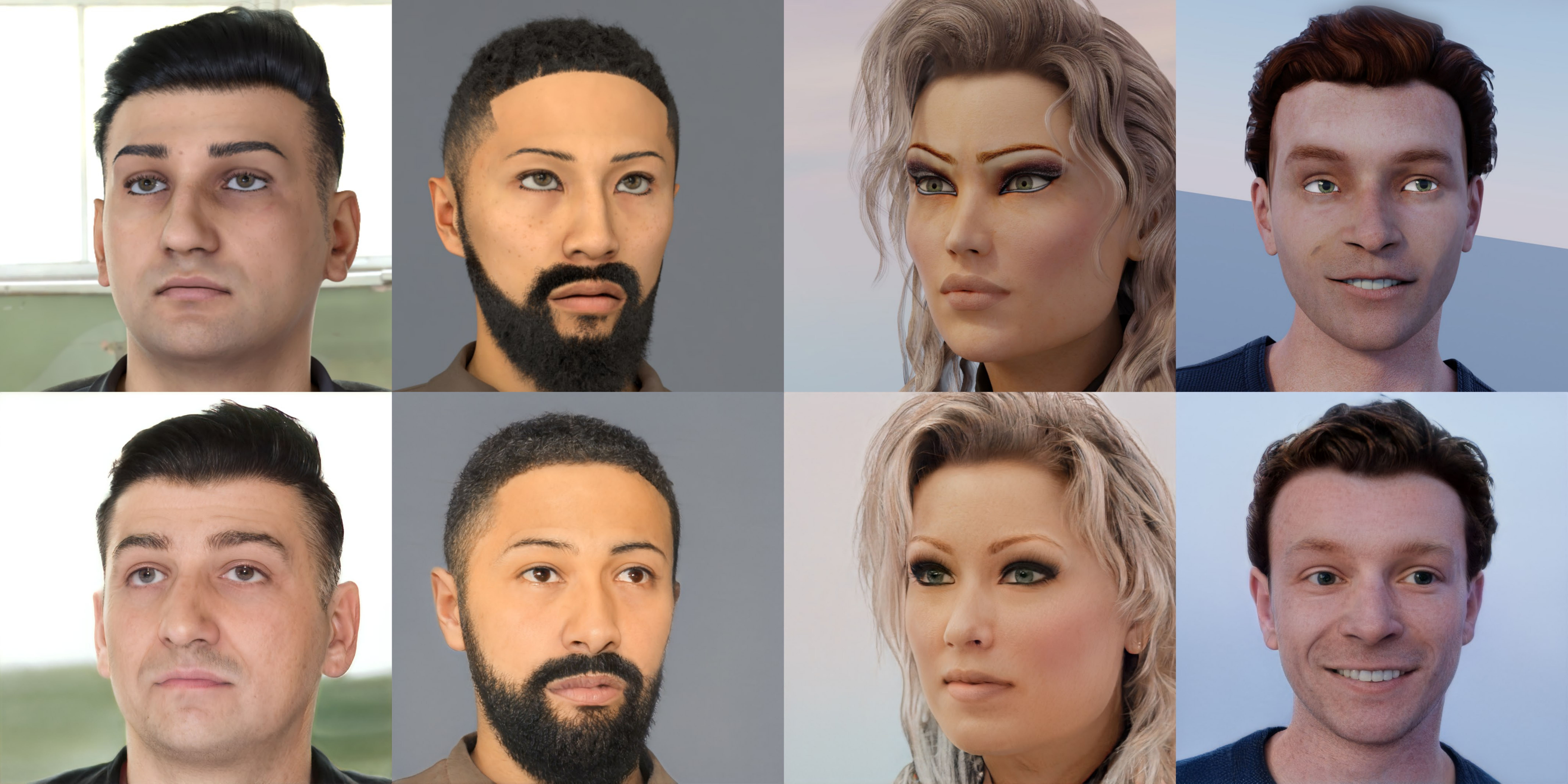}
\caption{\label{fig:results} Diverse photo-real faces generated by our method. The generated faces overcome the ``uncanny valley'' effect by maintaining identity, facial contour, and color.
The input rendered portraits are from the \textit{Diverse Human Faces} dataset (columns 1, 2) and Paul Schultz at Flickr \cite{Flickr} (columns 3, 4).}
\end{figure}

\section{Results}
\label{sec:Results}
{This section showcases the outcomes of our photo-realistic portrait generation framework}. 
We present the results of our approach as applied to a series of rendering-style portraits.
In Figs. \ref{fig:teaser} and \ref{fig:results}, we display a variety of results that span various genders, ages, and races, effectively illustrating how our approach can adapt across diverse data sources
(\textit{e.g. Diverse Human Faces} dataset \cite{DHFdataset} and internet images).
Additionally, we also showcase some examples where we stitch the generated realistic faces back onto the original apparel display renderings {(refer to Fig. \ref{fig:display} below and Fig. 7 in the supplementary material).}
Our generated realistic faces can easily blend in with the rendered garment and virtual avatar bodies with only minor post-processing (see the supplementary material). The adoption of our method can significantly enhance the overall authenticity of apparel display renderings.
In sum, our method effectively overcomes the ``uncanny valley'' effect (see also the user study) by largely improving the authenticity of rendered faces while avoiding portrait infringement liability due to using generated faces.
{Furthermore, it preserves the facial identity, aligning with the designer's preference.}

\begin{figure}[h] 
\centering
\includegraphics[width=\linewidth]{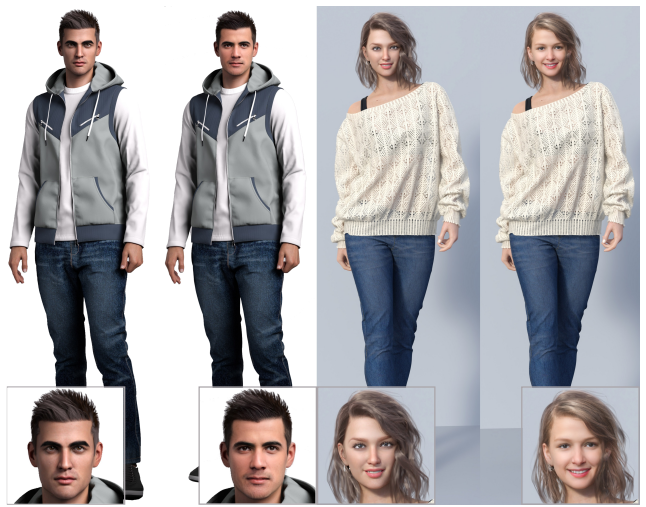}
\caption{\label{fig:display} We apply our method to digital apparel sample display images. Input images are courtesy of Yayat Punching at the CONNECT store \cite{CLO_connect}.}
\end{figure}

\begin{figure*}[htbp] 
\centering
\includegraphics[width=\linewidth]{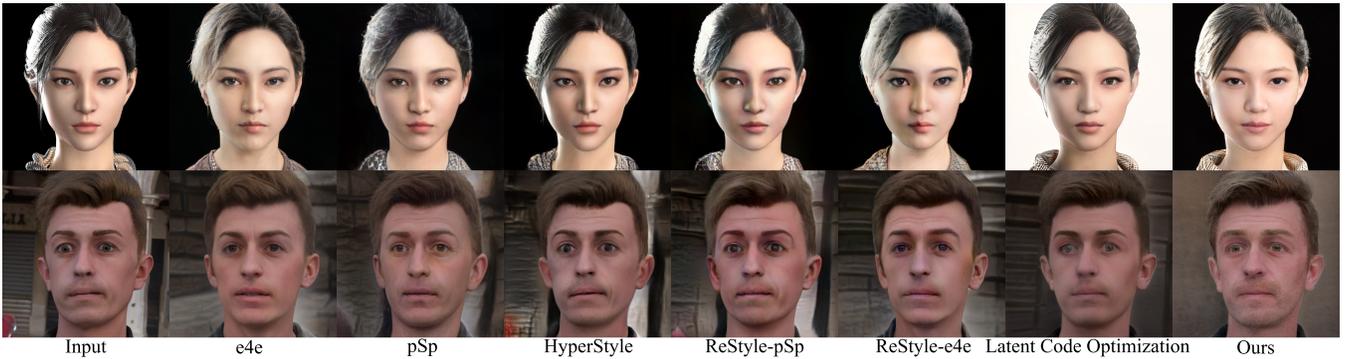}
\caption{\label{fig:realization_comparison} Qualitative comparisons with state-of-the-art StyleGAN inversion methods. From left to right, we show the input image, the results of 
e4e, pSp, HyperStyle, ReStyle-pSp, ReStyle-e4e, latent code optimization, and ours.
The testing input image in the bottom row is sourced from the \textit{Diverse Human Faces} \cite{DHFdataset} dataset.
Please zoom in for a better view.}
\end{figure*}

\section{Experiments}
\label{sec:Experiments}
In this section, we begin by presenting comparisons between our proposed method and state-of-the-art (SOTA) facial realism-improving methods (Sec. \ref{sec:realism-improving Experiments}). Subsequently, we provide comparisons between our identity-consistent style-transfer method and SOTA style-transfer methods (Sec. \ref{sec:style transfer Experiments}).

For an additional comprehensive understanding of our paper, please refer to the supplementary material. It provides in-depth information on ablation studies, {our method's application in digital sample display,}
implementation details concerning the network architecture and dataset, limitations and future work, and lightweight post-processing.

\subsection{Realism-improving}
\label{sec:realism-improving Experiments}
In this section, we conduct qualitative and quantitative experiments to demonstrate the effectiveness of our framework in improving facial realism. 
{The quantitative experiments are showcased through a user study.}

In Sec. \ref{sec::Related Work}, we mentioned that previous works, such as \cite{DBLP:conf/eccv/GarbinKJS20}, \cite{DBLP:journals/tog/ChandranWZRGGB21}, and some StyleGAN-inversion methods, can enhance the realism of rendered faces. However, the datasets and codes of \cite{DBLP:conf/eccv/GarbinKJS20} and \cite{DBLP:journals/tog/ChandranWZRGGB21} are not publicly accessible, so we rely on comparisons with state-of-the-art StyleGAN inversion methods.

\textbf{Qualitative experiments.}
To accomplish qualitative comparison, we directly project the input rendered portrait into the $W+$ latent space of StyleGAN2-\textit{FFHQ} using StyleGAN inversion methods, and then compare the inversion results with our own outcomes.
As illustrated in Fig. \ref{fig:realization_comparison}, we use e4e \cite{DBLP:journals/tog/TovANPC21}, pSp \cite{DBLP:conf/cvpr/RichardsonAPNAS21}, HyperStyle \cite{DBLP:conf/cvpr/AlalufTMGB22}, ReStyle \cite{DBLP:conf/iccv/AlalufPC21}, and latent code optimization \cite{10.1145/3544777}, for comparison.
Those encoders are trained on both \textit{FFHQ} dataset and StyleGAN2-\textit{FFHQ}.
For ReStyle, we run testing on both e4e and pSp encoders, using the ReStyle scheme.
For latent code optimization, we use the same inversion method described in Sec. \ref{sec:Inference} to project the input images into the $W+$ latent space of StyleGAN2-\textit{FFHQ}.
It is clear that those encoders lose many skin characteristics and produce faces with only smooth skin, which lacks realism. Furthermore, they retain the rendering style of the input images that looks unrealistic.
Our method, on the other hand, produces more photo-realistic results with more natural facial details and completely changes the input image's unrealistic rendering-style appearance while maintaining facial identity consistency.

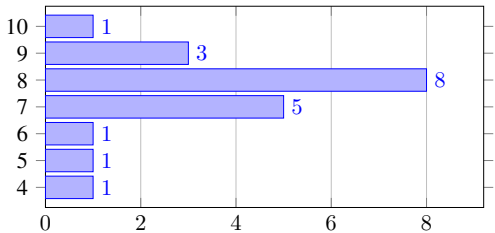
\begin{figure}[htbp]
\centering
\pgfplotstableread[col sep=comma]{user_study_0815_data.csv}\datacsv
\scalebox{0.85}{
\begin{tikzpicture}
    \begin{axis}[
        xbar,
        y=0.42cm, enlarge y limits={true, abs value=0.75},
        xmin=0, enlarge x limits={upper, value=0.15},
	xmajorgrids=true,
        ytick=data,
        yticklabels from table = {\datacsv}{a},
	nodes near coords, nodes near coords align=horizontal
    ]
    \addplot table [x=b, y=a]
    {\datacsv};
    \end{axis} 
\end{tikzpicture} 
}
\caption{\label{fig:user-study-realism}
The distribution of the user study on the authenticity comparison of the methods for improving facial realism. The $y-$axis shows the number of output portraits from our method chosen by participants (out of 10 sets), and the $x-$axis shows the number of participants. The results show that our method outperforms other methods for improving facial authenticity.}
\end{figure}

\textbf{User study.} 
To the best of our knowledge, there is no currently viable quantitative metric for assessing the authenticity of synthetic portraits. Furthermore, determining the authenticity of a portrait is largely dependent on human cognitive abilities. In light of this, we devised a user study as a quantitative experiment, with the goal of comparing the authenticity of the results produced by our proposed method to those produced by SOTA StyleGAN inversion methods. 
We collected ten rendered portraits and subjected them to the six StyleGAN inversion methods mentioned above (see qualitative experiments in Sec.~\ref{sec:realism-improving Experiments}) and our proposed method, respectively. We presented these ten sets of test cases sequentially to 20 participants, randomly displaying the results for authenticity comparison.
Fig.~\ref{fig:user-study-realism} shows that the vast majority of our results are more realistic. This demonstrates our approach's superiority over other StyleGAN inversion methods in improving the authenticity of rendered portraits.

\subsection{Style Transfer}
\label{sec:style transfer Experiments}

In this section, we conduct qualitative and quantitative experiments to demonstrate the effectiveness of our identity-consistent style transfer algorithm.  
We will show that our style-transfer approach surpasses other style transfer methods in both style transfer and facial identity preservation.

We compare our identity-consistent transfer method to the SOTA StyleGAN-based style transfer methods.
Since one-shot domain adaptation methods \cite{DBLP:conf/iclr/ZhuA0W22,Zhang2022GeneralizedOD} stylize the whole latent space using a single reference image, we cannot apply them to process our diverse testing images. 
Thus we make comparisons with StyleGAN-NADA \cite{DBLP:journals/tog/GalPMBCC22} and AgileGAN \cite{DBLP:journals/tog/SongLLMLZC21}. 

\begin{figure}[htbp] 
\centering
\includegraphics[width=\linewidth]{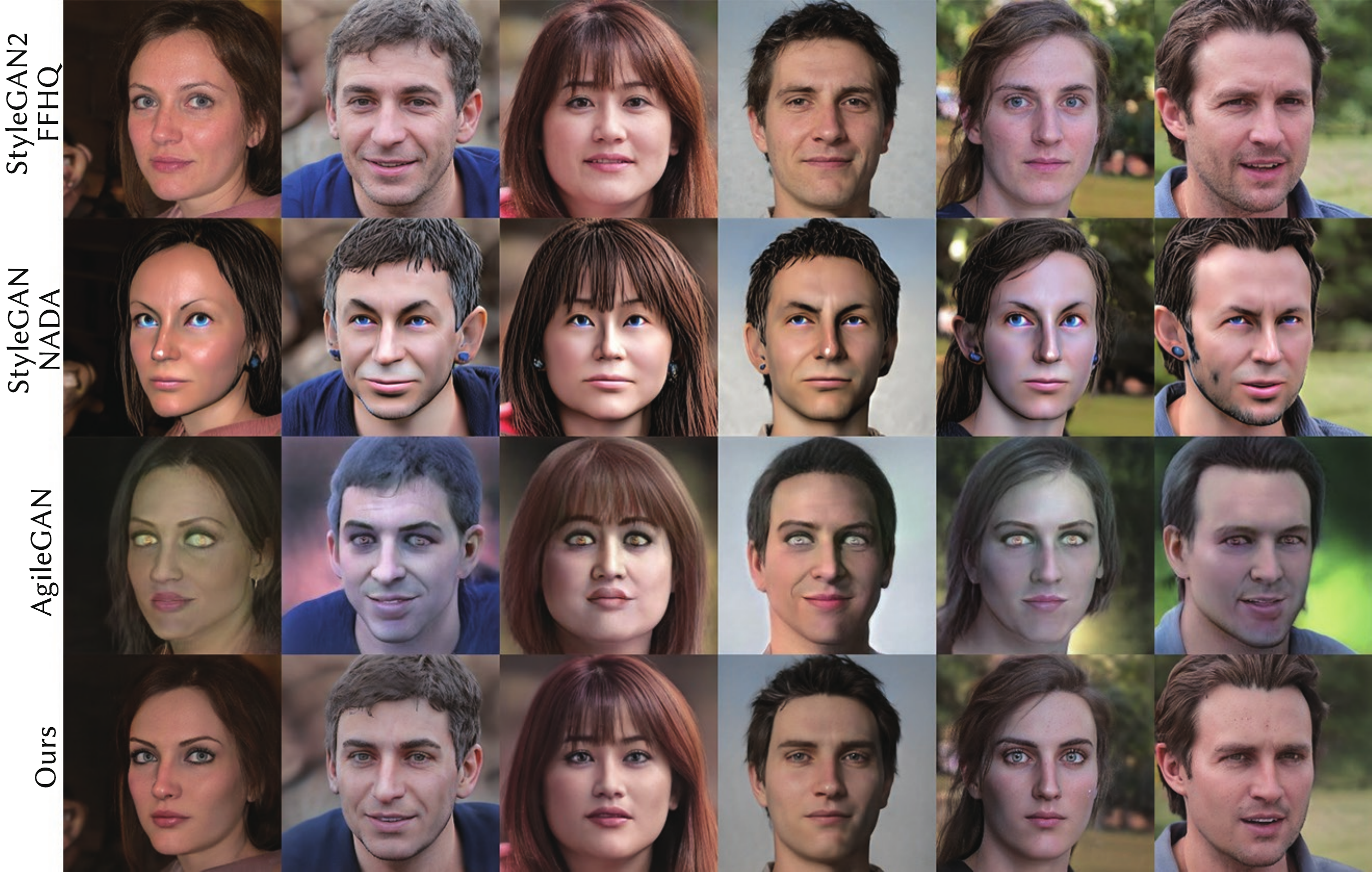}
\caption{\label{fig:stylization_comparison} Qualitative comparisons with state-of-the-art style transfer methods based on StyleGAN. 
From up to bottom, we present the images generated by StyleGAN2-FFHQ, StyleGAN-NADA, AgileGAN, and ours. Images in the same column are generated by the same latent code.}
\end{figure}

\textbf{Qualitative experiments.}
We present a comparison between the style transfer results of our identity-consistent style transfer method and those of StyleGAN-NADA \cite{DBLP:journals/tog/GalPMBCC22} and AgileGAN \cite{DBLP:journals/tog/SongLLMLZC21} in Fig. \ref{fig:stylization_comparison}.
For StyleGAN-NADA, we choose ``Photo'' as the source text and ``Rendered avatar'' as the target text.
For AgileGAN, we use our \textit{DRFHQ} dataset as the training dataset
to train AgileGAN.
We compare the images generated by different generators using the same latent code.
Results show that the StyleGAN-NADA semantic guidelines are too vague to produce acceptable results. AgileGAN generates artifacts and unnatural skin color. In contrast, our approach produces rendering-style results while preserving face identity.

\textbf{Fréchet Inception Distance metric.}
To evaluate the performance in transferring style to that of \textit{DRFHQ} dataset, we utilize Fréchet Inception Distance (FID) \cite{NIPS2017_FID} to measure the overall similarity between the distribution of synthesized images and that of the \textit{DRFHQ} dataset (see the 2nd column in Table \ref{tab:fid}). 
Besides, to evaluate the geometry and color preservation quality, we compute the FID of the synthesized rendering-style images with respect to the realistic-style \textit{FFHQ} dataset 
{(see the 3rd column in Table \ref{tab:fid}).
In Table \ref{tab:fid}, StyleGAN2-\textit{DRFHQ} represents our identity-consistent model finetuned on our \textit{DRFHQ} dataset, AgileGAN-\textit{DRFHQ} represents AgileGAN \cite{DBLP:journals/tog/SongLLMLZC21} finetuned on our \textit{DRFHQ} dataset. Since StyleGAN-NADA \cite{DBLP:journals/tog/GalPMBCC22} is text-guided and not trained on our \textit{DRFHQ} dataset, FID is for reference only. 
Our model achieves the lowest FID as shown in Table \ref{tab:fid}, indicating that our StyleGAN2-\textit{DRFHQ} model is better at {both} style transfer and facial identity preservation.

\textbf{Identity similarity metric.}
To further assess the performance in facial identity preservation, we utilize a pretrained CurricularFace network \cite{huang2020curricularface} to compute identity similarity during facial style transfer.
Specifically, we apply our StyleGAN2-\textit{DRFHQ} model, AgileGAN-\textit{DRFHQ} model, and StyleGAN-NADA model to convert the style of 2k images from realistic to rendering respectively, we then use the CurricularFace network to measure facial identity.
As shown in Table \ref{tab:identity-similarity}, our StyleGAN2-\textit{DRFHQ} model exhibits superior performance in preserving facial identity during the process of style transfer.

\begin{table}[t]
\footnotesize
\centering
\caption{Fréchet Inception Distances (FID) score for different StyleGAN-based style transfer methods and datasets, computed from randomly generated 50k images. Lower scores are better.}
\label{tab:fid}
\begin{tabular}{@{}ccc@{}}
\toprule
Algorithm                       & \textit{DRFHQ} & \textit{FFHQ} \\ \midrule
StyleGAN2-\textit{DRFHQ} (Ours) & \textbf{24.5}           &  \textbf{16.3 }        \\
AgileGAN-\textit{DRFHQ}         & 62.5           &  83.5         \\ 
StyleGAN-NADA                   & 49.9           &  53.8             \\ \bottomrule
\end{tabular}
\end{table}

\begin{table}[t]
\footnotesize
\centering
\caption{Identity similarity measurement for SOTA StyleGAN-based style transfer methods, computed from randomly generated 2k images. Higher scores are better.}
\label{tab:identity-similarity}
\begin{tabular}{cc}
\hline
Algorithm                       & Identity Similarity $\uparrow$ \\ \hline
AgileGAN-\textit{DRFHQ}         & 0.14                           \\
StyleGAN-NADA                   & 0.34                           \\
StyleGAN2-\textit{DRFHQ} (Ours) & \textbf{0.57}                  \\ \hline
\end{tabular}
\end{table}
\section{Conclusions}
\label{sec:Conclusions}
We present a novel identity-consistent transfer learning method that can effectively remove the rendering-style appearance in the input portraits and generate photo-realistic portraits. Besides, we create a high-quality rendering-style portrait dataset, Daz-Rendered-Faces-HQ (\textit{DRFHQ}), which includes 11,399 images with gender, age, pose, and race variations.
To maintain the facial identity, we employ sketch and color constraints in the finetuning process of the StyleGAN2 generator on the \textit{DRFHQ} dataset. 
During inference, we first leverage latent code optimization to the input rendering-style portrait, then feed the projected inversion latent code into the real-style StyleGAN2-FFHQ generator, and finally obtain the photo-realistic result with consistent identity. 
We apply our method to digital apparel sample display, and experiments show that it can improve the overall realism of digital apparel samples.
Moreover, our rendering-style \textit{DRFHQ} dataset has the potential to motivate other applications such as virtual avatar synthesis and editing.

\bibliography{aaai24}

\end{document}